\documentclass[runningheads]{llncs}
\pdfoutput=1 

\usepackage{graphicx}
\usepackage{comment}
\usepackage{amsmath,amssymb} 
\usepackage{color}
\usepackage{ulem}   

\usepackage{epsfig}
\usepackage{graphicx}
\usepackage{amsmath}
\usepackage{amssymb}
\usepackage[flushleft]{threeparttable}

\usepackage{amsfonts}
\usepackage{physics} 
\usepackage[protrusion=true,tracking=true,kerning=true,expansion,final]{microtype}      
\usepackage{xcolor} 
\usepackage{booktabs} 
\usepackage{subcaption}
\usepackage{nicefrac} 
\usepackage{cancel}
\usepackage[square,sort,comma,numbers]{natbib}
\usepackage{xspace}

\usepackage{gensymb}
\usepackage{bbding}
\usepackage{mathtools}
\usepackage{centernot} 
\usepackage{bigints}
\usepackage{dsfont} 
\usepackage{esint} 
\usepackage{authblk}
\usepackage{multirow}

\usepackage{algorithm}
\usepackage{algpseudocode}
\usepackage{algorithmicx}

\usepackage{varioref} 
\usepackage[pagebackref=true,breaklinks=true,letterpaper=true,colorlinks,bookmarks=false]{hyperref}
\usepackage[capitalise]{cleveref} 
\crefname{appsec}{appendix}{appendices}
\Crefname{appsec}{Appendix}{Appendices}

\usepackage{url}

\definecolor{mydarkblue}{rgb}{0,0.08,0.45}
\definecolor{urlcolor}{rgb}{0,.145,.698}
\definecolor{linkcolor}{rgb}{.71,0.21,0.01}
\hypersetup{ %
	pdftitle={Smoothed Inference for Adversarial Robustness}, 
	pdfauthor={Yaniv Nemcovsky, Evgenii Zheltonozhskii, Chaim Baskin, Brian Chmiel, Maxim Fishman, Alex M. Bronstein, Avi Mendelson},
	pdfsubject={},
	pdfkeywords={},
	pdfborder=0 0 0,
	pdfpagemode=UseNone,
	breaklinks=true,  
	colorlinks=true,
	letterpaper=true,
	bookmarks=false,
	urlcolor=urlcolor,
	linkcolor=linkcolor,
	citecolor=mydarkblue,
	filecolor=mydarkblue,
	pdfview=FitH}

\renewcommand*{\backref}[1]{} 
\renewcommand*{\backrefalt}[4]{%
	\ifcase #1 %
	\or
	(cited on p. #2)%
	\else
	(cited on pp. #2)%
	\fi
}
\makeatletter
\renewcommand{\@biblabel}[1]{#1.}
\makeatother

\vbadness=\maxdimen
\usepackage{bm}
\renewcommand{\vb}{\bm}
\renewcommand{\emph}{\textit}

\begin{document}
\pagestyle{headings}
\mainmatter
\def\ECCVSubNumber{7417}  

\title{Smoothed Inference for Improving Adversarial Robustness} 

\titlerunning{Smoothed Inference for Adversarial Robustness}
%

\newcommand*\samethanks[1][\value{footnote}]{\footnotemark[#1]}

\author{Yaniv Nemcovsky\thanks{Equal contribution.}\inst{1} \and Evgenii Zheltonozhskii\samethanks[1]\inst{1}\orcidID{0000-0002-5400-9321} \and Chaim Baskin\samethanks[1]\inst{1} \and Brian Chmiel\samethanks[1]\inst{1,2} \and Maxim Fishman\inst{1,2} \and Alex M. Bronstein\inst{1} \and Avi Mendelson\inst{1}}
\authorrunning{Y. Nemcovsky et al.}
%
\institute{Technion – Israel Institute of Technology \and
Intel – Artificial Intelligence Products Group (AIPG)
\email{ \href{mailto:yanemcovsky@cs.technion.ac.il}{yanemcovsky@cs.technion.ac.il};
\href{mailto:evgeniizh@campus.technion.ac.il}{evgeniizh@campus.technion.ac.il};
\href{mailto:chaimbaskin@cs.technion.ac.il}{chaimbaskin@cs.technion.ac.il};
\href{mailto:brian.chmiel@intel.com}{brian.chmiel@intel.com};
\href{mailto:bron@cs.technion.ac.il}{bron@cs.technion.ac.il};
\href{mailto:avi.mendelson@cs.technion.ac.il}{avi.mendelson@cs.technion.ac.il}}}
\maketitle

\begin{abstract}
	Deep neural networks are known to be vulnerable to adversarial attacks. 
	Current methods of defense from such attacks are based on either implicit or explicit regularization, e.g., adversarial training. Randomized smoothing, the averaging  of the classifier outputs over a random distribution centered in the sample, has been shown to guarantee the performance of a classifier subject to bounded perturbations of the input. In this work, we study the application of randomized smoothing as a way to improve performance on unperturbed data as well as to increase robustness to adversarial attacks. The proposed technique can be applied on top of any existing adversarial defense, but works particularly well with the randomized approaches. We examine its performance on common white-box (PGD) and black-box (transfer and NAttack) attacks on CIFAR-10 and CIFAR-100, substantially outperforming previous art for most scenarios and comparable on others. For example, we achieve 60.4\% accuracy under a PGD attack on CIFAR-10 using ResNet-20,  outperforming previous art by 11.7\%. Since our method is based on sampling, it lends itself well for trading-off between the model inference complexity and its performance. 
	A reference implementation of the proposed techniques is provided at \href{https://github.com/yanemcovsky/SIAM}{the paper repository}.
	
\end{abstract}


\section{Introduction}
\label{sec:intro}
Deep neural networks (DNNs) are showing spectacular performance in a variety of computer vision tasks, but at the same time are susceptible to adversarial examples -- small perturbations that alter the output of the network \cite{goodfellow2014explaining, szegedy2013intriguing}. Since the initial discovery of this phenomenon in 2013, increasingly stronger defenses
\cite{goodfellow2014explaining,madry2018towards,xie2019feature,sarkar2019enforcing,khoury2019adversarial,zhang2019defending,rakin2018parametric,DBLP:conf/icml/ZhangYJXGJ19} and counterattacks \cite{goodfellow2014explaining, carlini2017towards,athalye2018obfuscated,madry2018towards,rony2019decoupling, papernot2017practical,chen2017zoo,li2019nattack} were proposed in the literature. Adversarial attacks have also been shown to occur in tasks beyond image classification where they were first discovered: in real-life object recognition \cite{brown2017adversarial,xu2019evading,athalye2017synthesizing},
object detection  \cite{wang2019daedalus}, natural language processing \cite{gao2018black, chaturvedi2019exploring,jin2019bert}, reinforcement learning \cite{gleave2019adversarial}, speech-to-text \cite{carlini2018audio}, and point cloud classification \cite{xiang2019generating}, just to mention a few. 
Moreover, the adversarial attacks can be used to improve the performance of the DNNs on unperturbed data \cite{xie2019adversarial,gong2020maxup,sun2020robust}.  

Understanding the root cause of adversarial examples, how they are created and how  we can detect and prevent such attacks, is at the center of many research works. 
\citet{gilmer2018adversarial} argued that adversarial examples are an inevitable property of high-dimensional data manifolds rather than a weakness of specific models.   
In view of this, the true goal of an adversarial defense is not to get rid of adversarial examples, but rather to make their search hard. 

Current defense methods are based on either implicit or explicit regularization. Explicit regularization methods aim to increase the performance under adversarial attack by directly incorporating a suitable term into the loss of the network during training, 
usually by incorporating adversarial examples for the dataset used in the training process.
In contrast, implicit regularization methods that do not change the objective, such as variational dropout \cite{kingma2015variational}, seek to train the network to be robust against any perturbations without taking into account adversarial examples.
In particular, adding randomness to the network can be especially successful \cite{liu2018towards,bietti2018kernel,rakin2018parametric}, since information acquired from previous runs cannot be directly applied to a current run.  Another way to make use of randomness to improve classifier robustness is randomized smoothing \cite{cohen2019certified}: averaging of the outputs of the classifier over some random distribution centered in the input data point. 
The effects of these three approaches (explicit regularization, implicit regularization, and smoothing) do not necessarily line up with or contradict each other. Thus, one could use a combination of the three, when devising adversarial defenses.

Previous works that discuss randomized smoothing do so exclusively in the context of certified robustness \cite{cohen2019certified,salman2019provably}. In contrast, we consider smoothing as a viable method to increase both the performance and adversarial robustness of the model. We show this effect on top of adversarial regularization methods -- both implicit  \cite{rakin2018parametric,our2020cpni} and explicit \cite{DBLP:conf/icml/ZhangYJXGJ19}.
We discuss several smoothed inference methods, and ways to optimize a pre-trained adversarial model to improve the accuracy of the smoothed classifier. 

\paragraph{Contributions.} This paper makes the following contributions.
Firstly, we study the effect of randomized smoothing on empirical accuracy  of the classifiers, both on perturbed and clean data. We show that even for a small amount of samples, accuracy increases in both cases. In addition, since the performance grows with the sample size, smoothing introduces a trade-off between inference time complexity and accuracy.

Secondly, we show that the smoothing can be applied along with any adversarial defense, and that it is especially efficient for methods based on implicit regularization by noise injection, such as PNI \cite{rakin2018parametric}.

Lastly, we propose a new family of attacks based on smoothing and demonstrate their advantage  for adversarial training over conventional PGD.

The rest of the paper is organized as follows: \cref{sec:related} reviews the related work, \cref{sec:local} describes our proposed method,  \cref{sec:global} describes integration of the method with adversarial training, \cref{sec:exp} provides the experimental results, and \cref{sec:conclusion} concludes the paper.

\section{Related work}
\label{sec:related}
In this section, we briefly review the notions of white and black box attacks and describe the existing approaches for adversarial defense and certified defense.
\paragraph{Adversarial attacks.} 
Adversarial attacks were first proposed by \citet{szegedy2013intriguing}, who noted that it was possible to use the gradients of a neural network to discover small perturbations of the input that drastically change its output. Moreover, it is usually possible to change the prediction to a particular class, i.e., perform a \emph{targeted} attack.  It is common to divide adversarial attacks into two classes: \emph{white box} attacks, which have access to the internals of the model (in particular, its gradients); and \emph{black box} attacks, which have access only to the output of the model for a given input.

\paragraph{White box attacks.}
One of the first practical white box adversarial attacks is the fast gradient sign method (FGSM) \cite{goodfellow2014explaining}, which 
utilizes the (normalized) sign of the gradient as an adversarial perturbation:
\begin{align}
    \hat{\vb{x}} = \vb{x} + \epsilon \cdot \mathrm{sign}(\grad_{\vb{x}}\mathcal{L}),
\end{align}
where $\vb{x}$ and $\hat{\vb{x}}$ denote the clean and perturbed inputs, respectively, $\mathcal{L}$ is the loss function of the network, which the attacker tries to maximize, and $\epsilon$ is the desired attack strength.

\citet{madry2018towards} proposed using iterative optimization -- specifically, projected gradient ascent -- to find stronger adversarial examples:
\begin{align}
    \hat{\vb{x}}^k = \Pi_{B(\vb{x} , \epsilon)} \qty[\hat{\vb{x}}^{k-1} + \alpha \cdot \mathrm{sign}(\grad_{\vb{x}}\mathcal{L})].
\end{align}
The projection operator $\Pi$ restricts the perturbed input to be in some vicinity $B(\vb{x} , \epsilon)$ of the unperturbed input. The iterations are initialized with $\hat{\vb{x}}^0=\vb{x}$.
This attack, referred to as PGD in the literature, is one of the most powerful attacks known to date. \citet{gowal2019alternative} further improved it by running a targeted PGD over a set of classes and choosing the best performing example, showing a notable improvement in different settings.

C\&W \cite{carlini2017towards} is a family of attacks using other norm constraints, in particular, the $L_0$, $L_2$ and $L_\infty$ norms. They solve a minimization problem
\begin{align}
    \min &\norm{\vb{\delta}} + c \mathcal{L}(\vb{x} + \vb{\delta}).
\end{align}
In contrast to FGSM and PGD, which have a strict bound on the attack norm, the C\&W attack can work in unbounded settings, seeking a minimal perturbation that achieves the misclassification. 

\citet{rony2019decoupling} proposed to improve this method via decoupling direction and norm (DDN)  optimization, motivated by the fact that finding the adversarial example in a predefined region is a simpler task. The attack iteratively changes the norm depending on the success of a previous step:
\begin{align}
    \hat{\vb{x}}^k &= \Pi_{B(\vb{x}, \epsilon_k)} \qty[\hat{\vb{x}}^{k-1} + \alpha \cdot\grad_{\vb{x}}\mathcal{L}]\\
    \epsilon_k &= \qty(1 + s\cdot \gamma) \epsilon_{k-1},
\end{align}
where $s= -1$ if $\hat{\vb{x}}^{k-1}$ is misclassified and $s=1$ otherwise.

\paragraph{Black box attacks.}
The simplest way to attack a model $\mathcal{F}$ without accessing its gradients is to train a substitute model $\mathcal{F}'$ to predict the outputs of $\mathcal{F}$ \cite{papernot2017practical} and then use its gradients to apply any of the available white box attacks. \citet{liu2016delving} extended this idea to transferring the adversarial examples from one model (or ensemble of models) to another, not necessary distilled one from another. 

Other works proposed alternative methods for estimating the gradient. ZOO \cite{chen2017zoo} made a numerical estimation, NATTACK \cite{li2019nattack} uses natural evolution strategies \cite{wierstra2008natural}, and BayesOpt \cite{anonymous2020bayesopt} employed Gaussian processes. A detailed review of these strategies is beyond the scope of this paper. 

\paragraph{Adversarial defenses.} 
\citet{szegedy2013intriguing} proposed to generate adversarial examples during training and use them for training adversarially robust models, optimizing the loss
\begin{align}
    \mathcal{L}_{\text{adv}}(\vb{x},y) = (1-w) \cdot \mathcal{L}_{CE}(f(\vb{x}),y) + w \cdot \mathcal{L}_{CE}(f(\vu{x}),y), \label{eq:adv_loss}
\end{align}
where $\mathcal{L}_{CE}$ is the cross-entropy loss, $f$ is the classifier, $\vb{x}$ is a training instance with the label $y$, $\vu{x}$ is the corresponding adversarial example, and $w$ is a hyperparameter usually set to $w=0.5$.

This method is particularly convenient if the generation of adversarial examples is fast \cite{goodfellow2014explaining}. When combined with stronger attacks, it provides a powerful baseline for adversarial defenses \cite{madry2018towards}, and is often utilized as part of defense procedures. 


Many works proposed improvements over regular adversarial training by applying stronger attacks during the adversarial training phase. For example, \citet{khoury2019adversarial} proposed to use Voronoi cells instead of $\epsilon$-balls as a possible space for adversarial examples in the training phase. 
\citet{liu2019training} added adversarial noise to all the activations, not only to the input.
\citet{jiang2018learning} proposed using a learning-to-learn framework, training an additional DNN to generate adversarial examples, which is used to adversarially train the classifier, resembling GAN training. \citet{balaji2019instance} heuristically updated the per-image attack strength $\epsilon_i$, decreasing it if the attack succeeded and increasing otherwise. \citet{pmlr-v97-wang19i} proposed to gradually increase the strength of the attacks based on a  first-order stationary condition for constrained optimization.  

Randomization of the neural network can be a very powerful adversarial defense since, even if provided access to gradients, the attacker does not have access to the network, but rather some randomly perturbed version thereof. One of the first works involving randomization \cite{zheng2016improving}  proposed to improve robustness by 
 reducing the distance between two samples differing by a normally distributed variable with a small variance. \citet{zhang2019defending} proposed to add normal noise to the input, which was shown to reduce the Kullback-Leibler divergence between the clean and the adversarially-perturbed inputs. 
 
 TRADES \cite{DBLP:conf/icml/ZhangYJXGJ19} uses a batch of randomly perturbed inputs to better cover the neighbourhood of the point.  Together with replacing $\mathcal{L}_{CE}(f(\vu{x}),y)$ in \cref{eq:adv_loss} by $\mathcal{L}_{CE}(f(\vu{x}),f(\vb{x}))$, this provided a significant improvement over standard adversarial training. MART \cite{Wang2020Improving} further improves the defense performance by differentiating the misclassified and correctly classified examples.
 
Another random-based defense based on adversarial training is parametric noise injection (PNI) \cite{rakin2018parametric}. PNI improves network robustness by injecting Gaussian noise into parameters (weights or activations) with learned variance.
\citet{our2020cpni} extended the idea of PNI by proposing to inject low-rank multivariate Gaussian noise instead of independent noise. 

\citet{athalye2018obfuscated} observed that many defenses do not improve robustness of the defended network but rather obfuscate gradients, making gradient-based optimization methods less effective. They identified common properties of obfuscated gradients and organized them in a checklist. In addition, they proposed techniques to overcome common instances of obfuscated gradients: in particular, approximating non-differentiable functions with a differentiable substitute and using averaging on the randomized ones.

 

\emph{Randomized smoothing} 
\cite{cohen2019certified} is a method for increasing the robustness of a classifier by averaging its outputs over some random distribution in the input space centered at the input sample. \citet{cohen2019certified} has shown that randomized smoothing is useful for certification of the classifier \cite{wong2018provable} -- 
that is,  proving its performance under norm-bounded input perturbations. In particular, they have shown a tight bound on $L_2$ certification using randomized smoothing with normal distribution.
Consequently,  \citet{salman2019provably} used the smoothed classifier  to generate stronger ``smoothed'' adversarial attacks, and utilized them for adversarial training. Such training allows the generation of a more accurate base classifier, and as a result, improves  certified robustness properties.

\section{Randomized smoothing as an adversarial defense}
\label{sec:local}

A smooth classifier $\Tilde{f}_{\vb{\theta}}$ is a map assigning an input $\vb{x}$ the class label that the base classifier $f_{\vb{\theta}}$ is most likely to return for $\vb{x}$ under random perturbation $\eta$,
\begin{align}
    \Tilde{f}_{\vb{\theta}}(\vb{x}) &= \arg\max\limits_y P (f_{\vb{\theta}}(\vb{x}+\vb{\eta}) = y) = \arg\max\limits_y \int\limits_{\mathbb{R}^n} \dd{\vb{\eta}} p(\eta)\mathds{1}\qty[f_{\vb{\theta}}(\vb{x}+\vb{\eta}) = y], \label{eq:certified}
\end{align}
where $\vb{\eta}$ is some random vector and $p$ is its density function.
Since the integral (\ref{eq:certified}) is intractable, it should be approximated with the Monte Carlo method, i.e., by averaging over a number of points from the distribution sampled independently. We denote by $M$ the number of samples used for such an approximation.

In its most general form, the smoothed model output can be written as
\begin{align}
    \Tilde{f}_{\vb{\theta}}(\vb{x}) = \arg\max\limits_y \int\limits_{\mathbb{R}^n} \dd{\vb{\eta}} p_{\vb{\eta}}(\eta)V( p_{iy} (\vb{x}) ) \approx \arg \max\limits_y \sum_{i=1}^M V( p_{iy} (\vb{x}) ),
\end{align}
where $V$ is some voting function  and $p_{iy}(\vb{x})$ is the probability of class $y$ being predicted for the $i$-{th} sample $\vb{x}+\vb{\eta}_i$.

Since the smoothing is independent of the architecture of the base classifier $f_{\vb{\theta}}$, we can take any (robust) model and test the overall improvement provided by smoothing. 
In contrast to previous works that also employed randomized smoothing \cite{cohen2019certified,salman2019provably},  we use it to improve the empirical adversarial robustness rather than the certified one. While certification is a very strong and desired guarantee, researchers were unable to achieve practical $\ell_\infty$ certification radii and it is unclear whether this is possible at all \cite{blum2020random,yang2020randomized}. Nevertheless, smoothing  can still be employed as a practical method of improving empirical performance of the classifier.

In what follows, we describe three different implementations of smoothing, differing in their voting function $V$.

\paragraph{Prediction smoothing.}
The simplest possible way to use multiple predictions is to perform prediction voting, i.e., to output the most frequent prediction among the samples. In other words, we take $V$ to be the following indicator function:
\begin{align}
    V(p_{iy}) &= \mathds{1} \qty[\arg \max\limits_{y'} p_{iy'} = y],
\end{align}    
yielding the following smoothed classifier,
\begin{align}
    \Tilde{f}_{\vb{\theta}}(\vb{x}) &= \arg \max\limits_y \sum_{i=1}^M \mathds{1} \qty[\arg \max\limits_{y'} f_{y'}(\vb{x}+\vb{\eta}_i) ],
\end{align}
where $f_{y'}(\vb{x})$ denotes the predicted probability that input $\vb{x}$ will belong to class $y'$. 

This is the randomized smoothing previously discussed by \citet{cohen2019certified} and \citet{salman2019provably} for certified robustness. In \cref{sec:exp} we show that by taking into account $M$ predictions, the accuracy of the classifier increases on both clean and adversarially-perturbed inputs. 
It is important to emphasize that this voting scheme only takes into account the classification of each of the $M$ samples, discarding the predicted probabilities of each class.

\paragraph{Weighed smoothing.}
The former method can be generalized by assigning some weight to top-$k$ predictions. Let us denote by $k(p_{iy})$ the rank of class $y$ as predicted by $f_{\vb{\theta}}(\vb{x}+\vb{\eta}_i)$, i.e., $k=1$ for the most probable class, $k=2$ for the second most probable one, and so on. Then, for example, the following choices of $V$ are possible:
\begin{align}
    V(p_{iy}) &= 2^{1-k(p_{iy})},  \,\,\, \mathrm{or}\\
    V_C(p_{iy}) &= \begin{cases}
    1 & k(p_{iy})=1\\
    C & k(p_{iy})=2\\
    0 & \text{otherwise.}
    \end{cases}
\end{align}
In particular, $V_C$ expresses the dependency on the second prominent class noted by \citet{cohen2019certified}. 

\paragraph{Soft prediction smoothing.}
In this case, we calculate the expectation of probability for each class and use them for prediction generation:
\begin{align}
    \Tilde{f}_{\vb{\theta}}(\vb{x}) = \arg \max\limits_y \sum_{i=1}^M \mathrm{softmax}\qty( f_{y}(\vb{x}+\vb{\eta}_i) ).
\end{align}

This method was previously mentioned by \citet{salman2019provably} as a way to apply adversarial training to a smoothed classifier. Since the probabilities for each class are now differentiable, this allows  the classifier to be trained end-to-end.
In contrast to prediction smoothing, we now 
fully take into account the predicted class probabilities of each of the $M$ samples. 
If, however, used as an adversarial defense, soft smoothing is easier to overcome by attackers. Even if the attack is not successful, the probability of competing classes increases, which means that even an unsuccessful attack on a base model does affect the prediction of the smoothed model.  In \cref{subsec:soft_training} we describe how this kind of smoothing can be used to create stronger attacks that can be used for adversarial training.

\subsection{Relation to implicit regularization}
\label{subs:input_noise}
In the case of prediction smoothing (or any other case in which $V$ is not differentiable), we cannot train the smoothed model directly. We, therefore, would like to train the base model in a way that optimizes the loss of the smoothed model. 
To this end, we use the 0-1 loss of $n$ training samples
\begin{align}
&\mathcal{L}_{01} = \sum_{i=1}^n \ell_{01} (y_i, \tilde{f}_{\vb{\theta}}(\vb{x}_i)),
\end{align}
with the pointwise terms
\begin{align}
& \ell_{01} (y_i, \tilde{f}_{\vb{\theta}}(\vb{x}_i)) = 1 - \mathds{1}\qty[y_i = \arg\max\limits_{y\in \mathcal{Y}} P_{\eta} \qty\big[f_{\vb{\theta}} (\vb{x}_i + \vb{\eta}) = y]], 
\end{align}
minimized over the model parameters $\vb{\theta}$.
Denoting for brevity $P_y = P_{\eta} \qty\big[f_{\vb{\theta}}(\vb{x}_i + \vb{\eta}) = y]$,
we can approximate the indicator as
\begin{align}
    &\mathds{1}\qty[y_i = \arg\max\limits_{y\in \mathcal{Y}} P_y] =
    \mathds{1}\qty[P_{y_i} \geq \max_{y'\in \mathcal{Y}\setminus \qty{ y_i}} P_{y'}] \approx\\
    \approx& \mathrm{ReLU}\qty[ P_{y_i} - \max_{y'\in \mathcal{Y}\setminus \qty{ y_i}} P_{y'}] =
    \max_{y\in \mathcal{Y}}\qty[P_y - \max_{y'\in \mathcal{Y}\setminus \qty{ y_i}} P_{y'}],  \label{eq:01loss}
\end{align}
where we approximate the Heaviside function $\mathds{1}$ with a better-behaving ReLU function on the interval $[-1,1]$. The last equality follows from the fact that $y=y'$ unless $y_i$ is the most probable class.
The expression in \cref{eq:01loss} resembles the bound \citet{cohen2019certified} has suggested for the radius of certification under adversarial attacks.

We now show a relation to training the base model under perturbation, similarly denoting $\mathds{1}_y = \mathds{1}\qty[f_{\vb{\theta}}(\vb{x}_i + \vb{\eta}) = y]$:
\begin{align}
\ell_{01} (y_i, \Tilde{f}_{\vb{\theta}}(\vb{x}_i)) =
1 - 
\max_{y\in \mathcal{Y}} \mathbb{E}_{\eta} 
\qty[\mathds{1}_y - 
\max_{y'\in \mathcal{Y}\setminus \qty{ y_i}} 
\mathds{1}_{y'}].
\end{align}
Written in this form, the 0-1 loss is now amenable to Monte Carlo approximation; however, working with such a non-convex loss is problematic. We, therefore, bound the $\max$ over the expectation by the expectation over the $\max$,
\begin{align}
\ell_{01} (y_i, \Tilde{f}_{\vb{\theta}}(\vb{x}_i)) \le 
1 - 
\mathbb{E}_{\eta} \max_{y\in \mathcal{Y}}  
\mathds{1}_y + 
\mathbb{E}_{\eta} \max_{y'\in \mathcal{Y}\setminus \qty{ y_i}} 
\mathds{1}_{y'}.
\end{align}
Since the classification events are disjoint, we obtain
\begin{align}
\mathcal{L}_{01} &=  \mathbb{E}_{\eta}\qty[\sum_{i=1}^n \left( 1 - \sum_{y\in \mathcal{Y}}  \mathds{1}_y + \sum_{y'\in \mathcal{Y}\setminus \qty{ y_i}}  \mathds{1}_{y'} \right) ] =\\&= \mathbb{E}_{\vb{\eta}}\qty[\sum_{i=1}^n 1 - \mathds{1}_{y_i}] =
\mathbb{E}_{\vb{\eta}}\sum_{i=1}^n\ell_{01} (y_i, f_{\vb{\theta}}(\vb{x}_i)), \label{eq:l01_final}
\end{align}
which is the 0-1 loss of the base classifier under Gaussian perturbation.
In addition, for the $\ell$-th layer, we can rewrite the network inference as 
\begin{align}
    f(\vb{x}_i) = f_{2} \circ f_{1} (\vb{x}_i) = f_{2} (\vb{x}'_i),
\end{align}
where $f_{1}$ denotes the first $\ell-1$ layers and $f_2$ stands for the rest of the network. Repeating the computation for $f_2$ and $\vb{x}'_i$ instead of $f$ and $\vb{x}_i$ shows that implicit regularization in the form of injecting Gaussian noise in any layer is beneficial for the smoothed classifier.

\paragraph{Noise strength learning.}
Having discussed the importance of noise injection for our method, we note that choosing the variance of the noise is not a trivial task. The optimal noise is dependent on both the architecture of the network and the task. Therefore, while considering the combination of our method with implicit regularization, we allow the variance of the noise injected into the layers to be learned as in PNI \cite{rakin2018parametric}.

\begin{figure*}
 \centering
 \includegraphics[width=\linewidth]{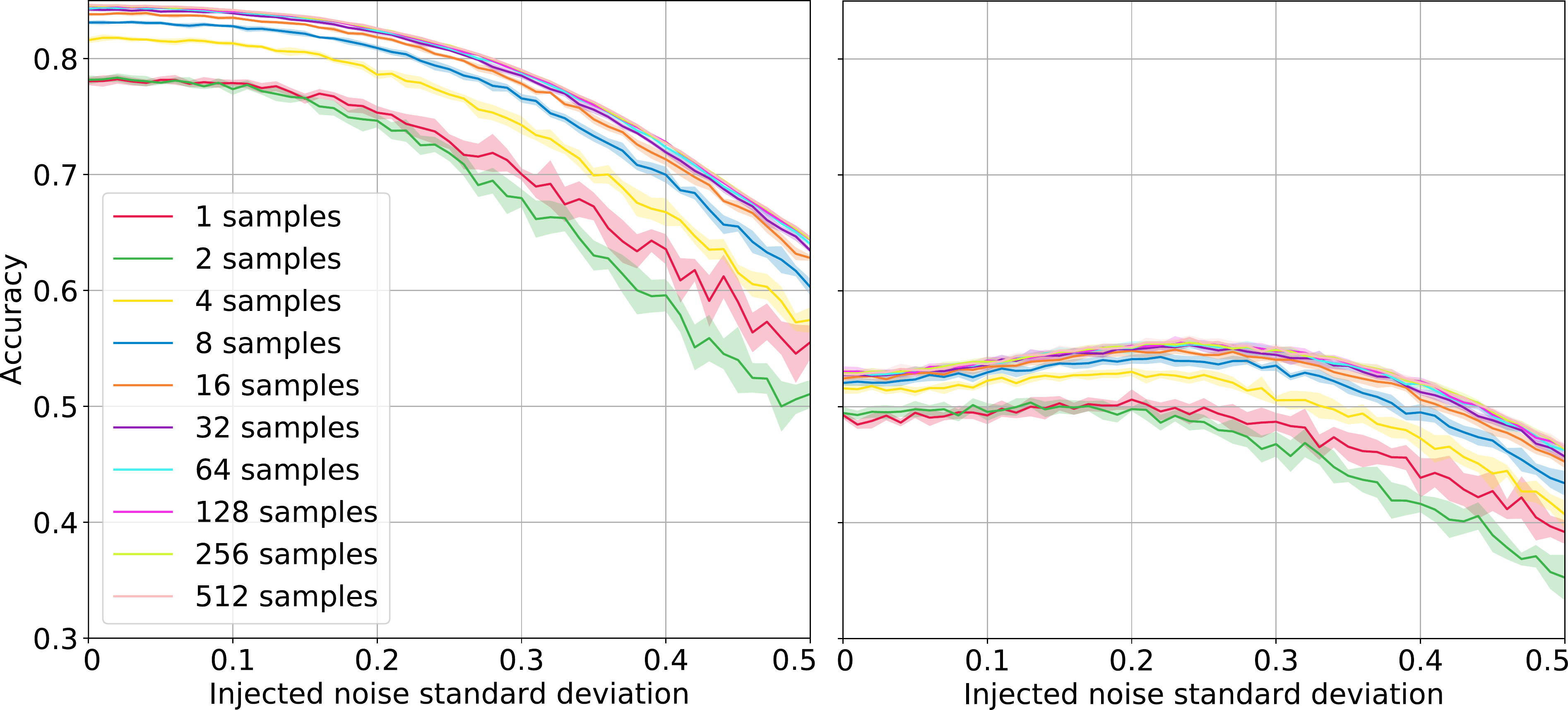}
 \caption{Accuracy as a function of injected noise strength with different numbers of samples for a CPNI base model with prediction smoothing on clean data (left) and under PGD attack (right). }
\label{fig:CPNI_model}
\end{figure*}

\section{Training a smoothed classifier}
\label{sec:global}
\citet{salman2019provably} have shown that incorporating the smoothing into the training procedure of the base classifier improves the aforementioned certification guarantees. We also use  a similar concept to improve the empirical accuracy of the smoothed classifier.
We consider two approaches to training the smoothed classifier: one based on implicit regularization and prediction smoothing, and another one based on soft prediction smoothing. In both cases, we start from an adversarially pre-trained base model and fine-tune it to improve the robustness of the smooth classifier.

\subsection{Prediction smoothing training}
\label{subsec:training}

From \cref{eq:l01_final} we conclude that by injecting Gaussian noise into   the base classifier layers,
we minimize the loss of the smoothed counterpart. The way the injected noise  propagates through the layers of the network, however, renders the noise injected into the input layer particularly important.
In addition, smoothing is inherently dependent on the strength of the noise injected during the process, and we show that similar models tend to achieve the best results around specific values of noise strength. We, therefore, fine-tune the classifier by training under noise injection of the same type and strength.

$\vb{\eta}$ is not restricted to the Gaussian distributions as long as the expectations in the derivations presented in the previous sections exist. One can furthermore combine Gaussian noise injection with adversarial training by letting
\begin{align}
\vb{\eta} & \sim \mathcal{N}(\vb{0}, \sigma^2\vb{I} ) + B \cdot \vb{\delta}  \nonumber\\
B &\sim \mathrm{Ber}(q),
\end{align}
where $\vb{\delta}$ is the adversarial attack  and $q$ is the weight of adversarial samples in the training procedure. Such adversarial training under Gaussian perturbations relates to minimizing the 0-1 loss of the smoothed classifier under adversarial attack. We refer to this method as CNI-I+W (input + weight noise).

\subsection{Soft smoothing training}
\label{subsec:soft_training}
The smoothing of the classifier gives rise to a new family of attacks targeting the smoothed classifier rather than the base model. In the case of soft smoothing, the output of the smoothed classifier is differentiable and thus can be used as a source of gradients for white box attacks. \citet{salman2019provably} discussed this family of attacks as a way to improve the generalization of a base model and thus improve certified accuracy of a smoothed model. 

We consider adversarial training with such attacks as a way to directly optimize the smoothed model by training the base model. We expect it to increase the adversarial robustness of both the base and the smoothed models. Specifically, we consider PGD attacks where the gradients are computed based on several randomized samples. In our experiments, we chose to limit the number of samples to $M=8$, since further improvement in the prediction ability is counter-balanced by the increase in the  computational complexity.

\paragraph{Expectation PGD.}
The aforementioned soft smoothing was previously explored in various ways. \citet{kaur2019perceptually} showed that targeted attacks on softly smoothed models have perceptually-aligned gradients, i.e., they perturb the input to make it resemble a member of a different class, even for models that were not adversarially trained. In contrast, non-smoothed classifiers exhibit this phenomenon only in adversarially-trained models. 
In general, perceptual alignment of the gradients is indirect evidence of model robustness \cite{ilyas2019adversarial,nakkiran2019a}. We expect such a perturbation, if strong enough, to be able to attack even a perfect classifier. Such phenomena might indicate that adversarial training of the model makes the adversarial attacks converge to soft smoothing-based attacks, which use the spatial information in proximity to the data point. 

An important difference between the  $\textsc{SmoothAdv}_{\mathrm{PGD}}$ attack considered by \citet{salman2019provably} and \citet{kaur2019perceptually}, and the \emph{expectation PGD} (EPGD) attack proposed in this paper is the order of the averaging and the softmax operator:
\begin{align} 
    g_{\text{SmoothAdv}}(\vb{x}) &=  \mathrm{softmax}\qty( \frac{1}{M} \sum_{i=1}^M  f_{c}(\vb{x} + \vb{\eta}_i) ) \,\,\, \mathrm{vs.} \\
    g_{\text{EPGD}}(\vb{x}) &=  \sum_{i=1}^M \mathrm{softmax}\qty( f_{c}(\vb{x} + \vb{\eta}_i) ).
\end{align}
Notice that an EPGD attack with $k$ PGD steps and $M$ samples requires computing the gradients $k\times M$ times, making its computational cost comparable with that of a PGD attack with $k\times M$ steps. Therefore, we compare the accuracy of PGD, EPGD and $\textsc{SmoothAdv}_{\mathrm{PGD}}$ \cite{salman2019provably} as a function of $k \times M$ and find that each has a distinct behavior. 

\section{Experiments}
\label{sec:exp} 
To demonstrate the effectiveness of the proposed method, we evaluated of the proposed adversarial defenses on CIFAR-10 and CIFAR-100 under white and black box attacks for ResNet20 and Wide-ResNet34-10. Finally, we performed an extensive ablation study of our method.

\paragraph{Experimental settings.}

We studied the performance of each of the proposed smoothing methods (prediction, soft prediction, and weighed smoothing) over different base models. We first considered implicit regularization methods and combined them with the adversarial training proposed by \citet{madry2018towards}. We trained five such models: adversarially-trained CNI-W (CNI with noise added to weights), and four additional models obtained by fine-tuning the base model with the training methods described in \cref{sec:global}. One of those models is CNI-I+W (CNI with noise added to weights and input), while the remaining ones were obtained by using different attacks during the fine-tuning. We used EPGD attacks, $\textsc{SmoothAdv}_{\mathrm{PGD}}$ attacks, or PGD with a large number of iterations $k$. For EPGD attcks, we selected models based on clean validation set performance: one with the best validation accuracy (labelled \emph{EPGD-converged}) and another with the worst validation accuracy (\emph{EPGD-diverged}). The motivation for considering the latter derives from the fact that PGD and EPGD are highly similar in nature, to the point that PGD converges to EPGD with adversarial training. Therefore, training until convergence may result in a model very similar to CNI-W. 

\begin{figure*}
 \centering
  \includegraphics[width=\linewidth]{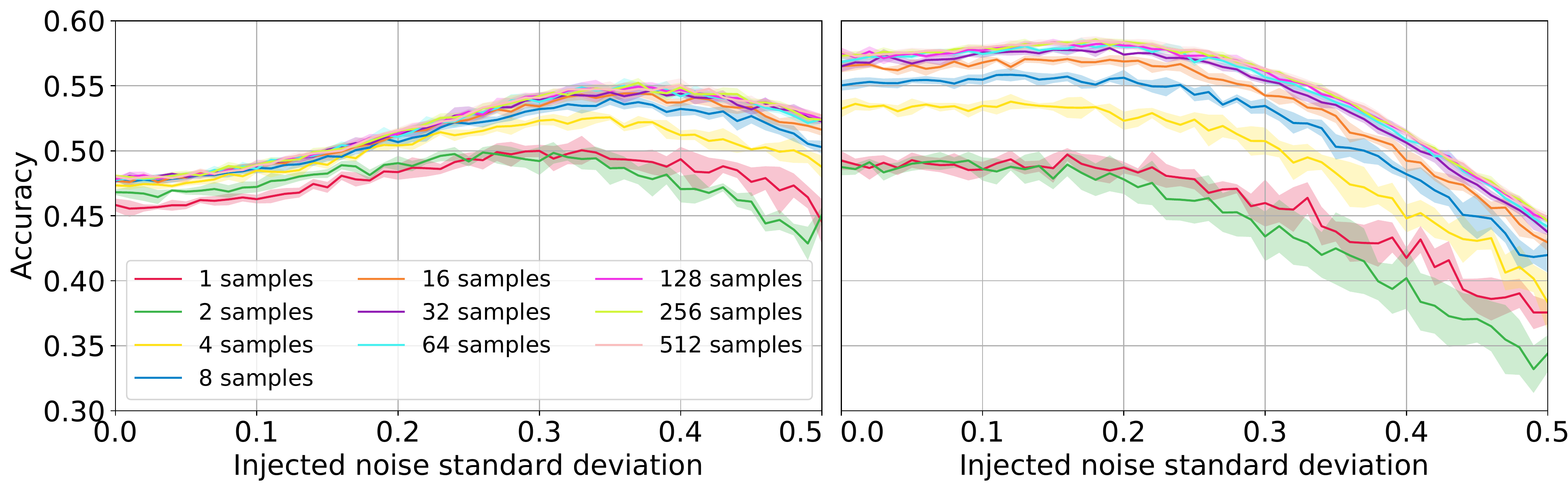}
 \caption{Accuracy as a function of injected noise strength with different numbers of samples for  CNI-I+W (left) and  EPGD-diverged (right) with prediction smoothing for PGD.
}
\label{fig:EPGD100_Gauss_model}
\end{figure*}

For the PGD attack we used $k=56$ to get  computational complexity similar to EPGD. Since this attack is too strong, the base model achieved significantly worse performance on both clean and adversarial data. Nevertheless, after the application of smoothing, this model  showed the highest performance on adversarial data 
with a somewhat lower accuracy on clean data. Secondly, we used TRADES (model denoted CNI-W+T) to combine both implicit and explicit regularization in a single model, expecting to improve on each of those separately.
For CNI-I+W,  we fine-tuned the CNI-W model based on the noise maximizing performance, $\sigma = 0.24$.
For $\textsc{SmoothAdv}_{\mathrm{PGD}}$, we used the method described in \cite{salman2019provably}. We report a number of the results for each base model and smoothing method  in \cref{tab:results_wb_cf}, and a comparison of different setups in \cref{fig:CPNI_model,fig:EPGD100_Gauss_model}. 
From \cref{fig:EPGD100_Gauss_model} we conclude that the \emph{EPGD-converged} and CNI-W base models result in similar performance for all setups. 

For  ResNet-20 on CIFAR-10, we used the CNI-W base model \cite{our2020cpni} trained for 400 epochs and chose the model with the highest performance on a validation set. 
The obtained model was fine-tuned with EPGD adversarial training, CNI-I+W training or CNI-W with a higher $k$ for up to $200$ epochs. For Wide-ResNet34-10 on CIFAR-100,  we used the PNI-W model \cite{rakin2018parametric} trained for 100 epochs and chose the model with the highest performance on a clean validation set. For Wide-ResNet34-10 on CIFAR-10, we used the same training regime, additionally using the PNI-W+Trades model that was obtained by training a PNI-W architecture with the TRADES method instead of adversarial training. 
\subsection{Comparison to other adversarial defenses}
\begin{table}
    \centering
    \caption{
    Comparison of our method to prior art on CIFAR-10 on ResNet-20, under PGD attack with the number of iterations $k=7$ and the $\ell_\infty$ radius of attack $\epsilon = \nicefrac{8}{255}$.  $^\dagger$ denotes our results based on code provided by the authors or our re-implementation.
    Smooth CNI-W uses $k=7$ and $M=512$ with $\sigma=0.24$ for the regular setting  and $\sigma=0.00$ for the no-noise setting during inference. For the high $k$ setting, $k=56$ was used for adversarial training and $M=512$,  $\sigma=0.01$ during inference. 
  }
    \centering
    \begin{tabular}{@{}lcc@{}} 
        \toprule
        \multirow{2}{*}{\textbf{Method}}&\multicolumn{2}{c}{\textbf{Accuracy,  mean$\bm{\pm}$std\%} } \\ 
        \cmidrule(r){2-3}
         & \textbf{Clean} & \textbf{PGD}   \\ 
        \midrule
        Adversarial training \cite{madry2018towards}$^\dagger$   & $83.84$    & $39.14\pm0.05$ \\        
        PNI \cite{rakin2018parametric}   & $82.84\pm0.22$   & $46.11\pm0.43$ \\
        TRADES ($\nicefrac{1}{\lambda} = 1$) \cite{DBLP:conf/icml/ZhangYJXGJ19}$^\dagger$ & 75.62 & 47.18 \\
        TRADES ($\nicefrac{1}{\lambda} = 6$) \cite{DBLP:conf/icml/ZhangYJXGJ19}$^\dagger$ & 75.47  & 47.75 \\
        CNI \cite{our2020cpni}   & $78.48\pm0.41$    & $48.84\pm0.55$  \\
        Smoothed CNI-W (ours)  & $81.44\pm 0.06$    & $55.92\pm0.22$  \\
        Smoothed CNI-W  (no noise, ours) & $\mathbf{84.63\pm0.05}$ &$52.8\pm0.2$  \\
        Smoothed CNI-W (high $k$, ours)  & $78.10\pm0.12$    & $\mathbf{60.06\pm0.29}$  \\ 
        \bottomrule
    \end{tabular}
    \label{tab:comp_cifar10}
\end{table}

\begin{table}
    \centering
    \caption{
    Comparison of our method to prior art on CIFAR-10 (a) and CIFAR-100 (b) on Wide-ResNet-34, under PGD attack with $k=10$ and $\epsilon = \nicefrac{8}{255}$. $^\dagger$ denotes our results based on code provided by the authors or our re-implementation. $^+$ denotes our results based on the checkpoint provided by the authors. For CIFAR-10 smooth PNI-W+T and PNI-W, we use $k=10$ and $M=64$ with $\sigma=0.17$ for the regular setting and $\sigma=0.03$ for the low noise setting during inference. For CIFAR-100, we use $k=10$ and $M=16$ with $\sigma=0.05$.
     }
    \centering
        \begin{subfigure}{0.495\linewidth}
        \begin{tabular}{@{}lcc@{}} 
        \toprule
        \multirow{2}{*}{\textbf{Method}}&\multicolumn{2}{c}{\textbf{Accuracy, \%} } \\ 
        \cmidrule(r){2-3}
         & \textbf{Clean} & \textbf{PGD}  \\ 
        \midrule
        Adv. training \cite{madry2018towards}$^\dagger$    & $89.68$    & $46.61$ \\        
        PNI \cite{rakin2018parametric}$^\dagger$   & $89.26$   & $52.9$ \\
        IAAT \cite{balaji2019instance}  &$\mathbf{91.3}$ & $48.53$  \\
        CSAT \cite{sarkar2019enforcing} & $87.65$ & $54.77$ \\
        TRADES \cite{DBLP:conf/icml/ZhangYJXGJ19}$^+$ & $84.92$  &$56.5$   \\
        MART \cite{Wang2020Improving}$^+$ &$83.62$ &$\mathbf{57.3}$ \\
        Smooth PNI-W (our)  & $89.27$    & $54.73$  \\
        Smooth PNI-W+T  & \multirow{2}{*}{${85.71}$}    & \multirow{2}{*}{$56.82$}   \\
         ($\nicefrac{1}{\lambda} = 6$, our)  &     &  \\
        \bottomrule
    \end{tabular} 
    \subcaption{}
    \label{tab:comp_cifar10_wide}
    \end{subfigure}
    \hfill
    \begin{subfigure}{0.495\linewidth}
    \begin{tabular}{@{}lcc@{}} 
        \toprule
        \multirow{2}{*}{\textbf{Method}}&\multicolumn{2}{c}{\textbf{Accuracy, \%} } \\ 
        \cmidrule(r){2-3}
         & \textbf{Clean} & \textbf{PGD}  \\ 
        \midrule
        IAAT \cite{balaji2019instance}  & $\mathbf{68.1}$ & $26.17$  \\
        Adv. training \cite{madry2018towards}$^\dagger$  &  $65.8$  & $26.15$ \\
        PNI \cite{rakin2018parametric}$^\dagger$   & $61.72$   & $27.33$  \\
        Smooth PNI-W (our)  & $ 66.30 $    & $\mathbf{29.68}$  \\
        \bottomrule
    \end{tabular}
    \subcaption{}
    \label{tab:comp_cifar100}
    \end{subfigure}
\end{table}

For ResNet-20 on CIFAR-10, we compared our best-performing instance of the defense (smoothed CNI-W fine-tuned with a high $k$) to the current state-of-the-art. As summarized in \cref{tab:comp_cifar10}, in terms of adversarial accuracy, we outperformed the best existing method by 11.7\%. In addition, we presented a configuration of smoothed classifiers that achieved the highest accuracy on clean data, i.e., CNI-W without noise. \cref{tab:comp_cifar10_wide} presents the results of our method for Wide-ResNet-34 on CIFAR-10 against PGD with $k=10$ steps. 
Our smoothed PNI-W+T shows comparable results in terms of adversarial accuracy compared to the previous state-of-the-art. For Wide-ResNet-34 on CIFAR-100, our smoothed PNI-W substantially outperforms prior art in terms of adversarial accuracy, as shown in \cref{tab:comp_cifar100}.

       

   
    

\begin{table}
    \centering
    \caption{
    Results of black-box attacks on our method applied on PNI-W and CNI-W with prediction smoothing: (a) transferable attacks with $\sigma=0.02$;
    (b) NAttack. Our Smooth PNI-W and CNI-W method uses $M=1,4,8$, $\sigma=0.24$. $^\dagger$ denotes our results based on code provided by authors or our re-implementation. In both cases we present results for ResNet-20 on CIFAR-10
     }
    \centering
        \begin{subfigure}{0.47\linewidth}
    \begin{tabular}{@{}ccc@{}} 
        \toprule
        \multirow{2}{*}{\textbf{Iterations}}   &\multicolumn{2}{c}{\textbf{Accuracy,  mean$\bm{\pm}$std }} \\ 
        \cmidrule(r){2-3}
        & \textbf{PGD} & \textbf{PGD-s} \\ 
        \midrule
         4 &  $58.01\pm0.35$  & $59.35\pm0.15$   \\
          8 &$58.49\pm0.10$  & $60.23\pm0.14$  \\
        \bottomrule
    \end{tabular}
    \subcaption{}
    \label{tab:results_bb_cf}
    \end{subfigure}
    \hfill
    \begin{subfigure}{0.52\linewidth}
    \begin{tabular}{@{}cc@{}} 
        \toprule
         \textbf{Method} & \textbf{Accuracy, \%}  \\ 
        \midrule
        Adv. training \cite{madry2018towards}$^\dagger$  & $33.11$ \\
        Smooth PNI-W (our, $m=1$) &$47.17$\\
        Smooth PNI-W (our, $m=4$) &$50.29$\\
        Smooth PNI-W (our, $m=8$) &$50.78$\\
        Smooth CNI-W (our, $m=1$) & $48.83$  \\
        Smooth CNI-W (our, $m=4$) & $50.98$  \\
        Smooth CNI-W (our, $m=8$) & $\mathbf{51.56}$  \\
        \bottomrule
    \end{tabular}
    \subcaption{}
    \label{tab:Nattack}
    \end{subfigure}
\end{table}

We tested our defense against black-box attacks, in particular, the transferable attack  \cite{liu2016delving} and NAttack\cite{li2019nattack}. For the transferable attack, we trained another instance of the CNI-W model and used it as a source model in two configurations: PGD with and without smoothing. The results are reported in \cref{tab:results_bb_cf,tab:Nattack}. For the transferable attack, we conclude that our model performs well even if the source model is not smoothed, which is an argument against a randomized gradient obfuscation effect. The NAttack comparison shows that our method substantially outperforms the baseline benchmarks.

\subsection{Ablation study}

We compared a different number of Monte Carlo samples: $M=2^n$, $n\in \{0,\dots,9\}$. 
For each value of $M$, we considered multiple noise standard deviations in the range $[0,0.5]$. The upper bound that we chose is the result of significant degradation of the model's performance. 
 The best results were obtained with prediction smoothing, even for a low number of samples. Although the advantage is of the order of a single standard deviation, this difference is persistent across base models, number of iterations, and attack types, and thus is likely to be systematic.  This could indicate that the attackers utilize additional information provided by other methods better than the defender.


\begin{table*}
    \centering
    \caption{Results of Smooth CNI-W  for the PGD white-box attacks  on CIFAR-10. Mean and standard deviation over five runs is presented in the form of mean$\pm$std. }
    \begin{tabular}{@{}llcccc@{}} 
        \toprule
        \multirow{2}{*}{\textbf{Base Model}} & \multirow{2}{*}{\textbf{Smoothing}} & \multirow{2}{*}{\textbf{Iterations}} & \multirow{2}{*}{\textbf{Noise}} &\multicolumn{2}{c}{\textbf{Accuracy,  mean$\bm{\pm}$std }} \\ 
        \cmidrule(r){5-6}
         &&&& \textbf{Clean} & \textbf{PGD}  \\ 
        \midrule
        CNI-W  & Prediction  & 512 & 0.0 & $\mathbf{84.63\pm0.05}$  & $52.8\pm0.29$\\  
        CNI-W  & Prediction  & 512 & 0.24 & $81.44\pm0.06$  & $55.92\pm0.22$ \\ 
        EPGD-diverged  & Prediction  & 256 & 0.0 & $83.05\pm0.01$  & $57.31\pm0.22$\\ 
        EPGD-diverged   & Prediction  & 256 & 0.18 & $81.07\pm0.01$  & $58.56\pm0.14$ \\ 
        EPGD-diverged  & Soft  & 512 & 0.0 & $83.13\pm0.07$  & $57.07\pm0.15$  \\
        EPGD-diverged   & Soft  & 512 & 0.19 & $80.98\pm0.07$  & $58.32\pm0.33$ \\
        EPGD-diverged  & Weighted  & 256 & 0.19 & $80.98\pm0.07$  & $58.47\pm0.18$  \\
        EPGD-converged  & Prediction  & 256 & 0.22 & $81.82\pm0.05$  & $55.9\pm0.37$ \\ 
        CNI-I+W  & Prediction  & 256 & 0.37 & $81.25\pm0.06$  & $55.23\pm0.37$\\
        CNI-W ($k=56$) & Prediction  & 512 & 0.11 & $77.28\pm0.05$    & $\mathbf{60.54\pm0.26}$ \\
        \bottomrule
    \end{tabular}
    \label{tab:results_wb_cf}
\end{table*}

For the CNI-W and EPGD models, accuracy on clean data drops as the noise variance increases.
The CNI-I+W  model is more resilient to noise injection, acquiring maximum at around 40\% higher standard deviation of the injected noise.
This can be related to the fact that the certified radius is twice as small as the standard deviation of the noise used in the smoothed classifier. 

In contrast to the certified robustness scheme \cite{cohen2019certified}, we did not observe improvement of the CNI-I+W model over CNI-W, which might be a result of the interference of the CNI-induced noise and the Gaussian noise.

\section{Conclusions}
\label{sec:conclusion}
We proposed an approach to adversarial defense based on randomized smoothing, which shows state-of-the-art results for white-box attacks, namely PGD, on CIFAR-10 and CIFAR-100,  even with a small number of iterations. We also confirmed the efficiency of our defense against black-box attacks, by successfully defending against transferring adversarial examples from different models. 
Our method offers a practical trade-off between the inference time and model performance, and can be incorporated into any adversarial defense. We showed that adversarial training of a smoothed classifier is a non-trivial task and studied several approaches to it. In addition, we investigated a family of attacks that take smoothing into account against smoothed classifiers. By incorporating those attacks into adversarial training, we were able to train classifiers with higher performance in smoothed settings. 
Complexity-wise, however, when taking into account the computations required for smoothing during the training, they are comparable to but not better than other known attacks.
Finally, we show that our method can exploit both implicit and explicit regularization, which emphasizes the importance of incorporating implicit regularization, explicit regularization and smoothed inference together into adversarial defenses.

\subsubsection*{Acknowledgments}
The research was funded by the Hyundai Motor Company through the HYUNDAI-TECHNION-KAIST Consortium, National Cyber Security Authority, and the Hiroshi Fujiwara Technion Cyber Security Research Center.

\clearpage
%
%
\bibliographystyle{splncs04}
\bibliography{smoothed_eccv}

\newpage

\appendix{}

\renewcommand\thefigure{\thesection.\arabic{figure}} 
\renewcommand\thetable{\thesection.\arabic{table}} 
\renewcommand\theequation{\thesection.\arabic{equation}}  
\setcounter{figure}{0}  
\setcounter{table}{0}

\crefalias{section}{appsec}
\crefalias{subsection}{appsec}
\crefalias{subsubsection}{appsec}

\section{Defense by random noise injection }
To get some intuition on the effect of noise injection on adversarial attacks, we perform an analysis of a simple classification model -- a support vector machine (SVM) $f(\vb{x}) = \vb{w} \vdot \vb{x} + b$ with the parameters $\vb{w}$ and $b$. Specifically, we consider the formulation of SVM under adversarial attacks with zero-mean random noise injected into the input. We assume that the attacker is aware of this noise, but is limited to observing the effect of a single realization thereof.

We start from the expectation of the SVM objective on a single input sample $\vb{x}_i$ with ground truth label $y_i$:
\begin{align}
    f(\vb{x}_i) = \mathbb{E}_{\vb{\eta}} \max_{\vb{\delta}} \mathrm{ReLU}\qty\bigg[1-y_i (\vb{w} \vdot \qty(\vb{x}_i + \vb{\eta}_i - \vb{\delta}_i)  + b)], \label{eq:svm_expect}
\end{align}
where $\vb{\eta}_i$ is the injected noise and $\vb{\delta}_i$ is the adversarial noise. 
Denoting $\vb{\delta}'_i = \vb{\delta}_i - \vb{\eta}_i$, and using the result of \citet{xu2009robustness}, we write \cref{eq:svm_expect} as
\begin{align}
    f(\vb{x}_i) = \mathbb{E}_{\vb{\eta}} \max_{\vb{\delta}} \vb{w} \vdot \vb{\delta}'_i + \mathrm{ReLU}\qty\bigg[1-y_i (\vb{w} \vdot \vb{x}_i  + b)]
\end{align}
Since the expectation of $\vb{\eta}$ is $0$, $\mathbb{E}_{\vb{\eta}} \qty[\vb{w}\vdot \vb{\eta}_i ]= 0$, leading to
\begin{align}
    f(\vb{x}_i) =\max_{\vb{\delta}} \vb{w} \vdot \vb{\delta}_i + \mathrm{ReLU}\qty\bigg[1-y_i (\vb{w} \vdot \vb{x}_i  + b)],
\end{align}
which is nothing but the SVM objective without the injected noise.
Thus, the expectation of 
an adversarial attack 
will not change due to the noise injection, and it is unclear whether the attacker can devise a better strategy to breach the defense effect provided by the noise injection.

The effect of noise injection is similar to that of random gradient obfuscation. Let $\vb{x}$ be some input point, $\vb{\eta}$ a realization of the random perturbation, $\vb{\delta}$ the adversarial attack that would have been chosen for $\vb{x}$, and $\vb{\delta}'$ the adversarial attack corresponding to $\vb{x} + \vb{\eta}$. Since we add noise to the input in each forward iteration, the adversary computes $\vb{\delta}'$ instead of $\vb{\delta}$, which has some random distribution around the true adversarial direction. Denoting by $\Pi_a$ the projection on the direction chosen, by $\Pi_\perp$ the projection on the space orthogonal to this direction, and by $\norm{\vb{\delta}}_p \le \epsilon$ the $L_p$-bound on the attack strength, yields
\begin{align}
\Pi_\perp(\vb{\delta}') &= \Pi_\perp(\vb{\eta}) \equiv \vb{\eta_0}\\
\norm{\Pi_a\qty(\vb{\delta}')}_p &= \qty(\epsilon^p - \norm{\vb{\eta_0}}_p^p)^{\nicefrac{1}{p}}
\end{align}
For $p=\infty$, the second term equals $\epsilon$. The first term, however, is a random variable that moves $\vb{\delta}'$ farther away from the adversarial direction and, therefore, decreases the probability of a successful adversarial attack. This effect accumulates when the adversary computes the gradients multiple times (such as in PGD). 

\section{Additional results}

\begin{figure*}
 \centering
\includegraphics[width=\linewidth]{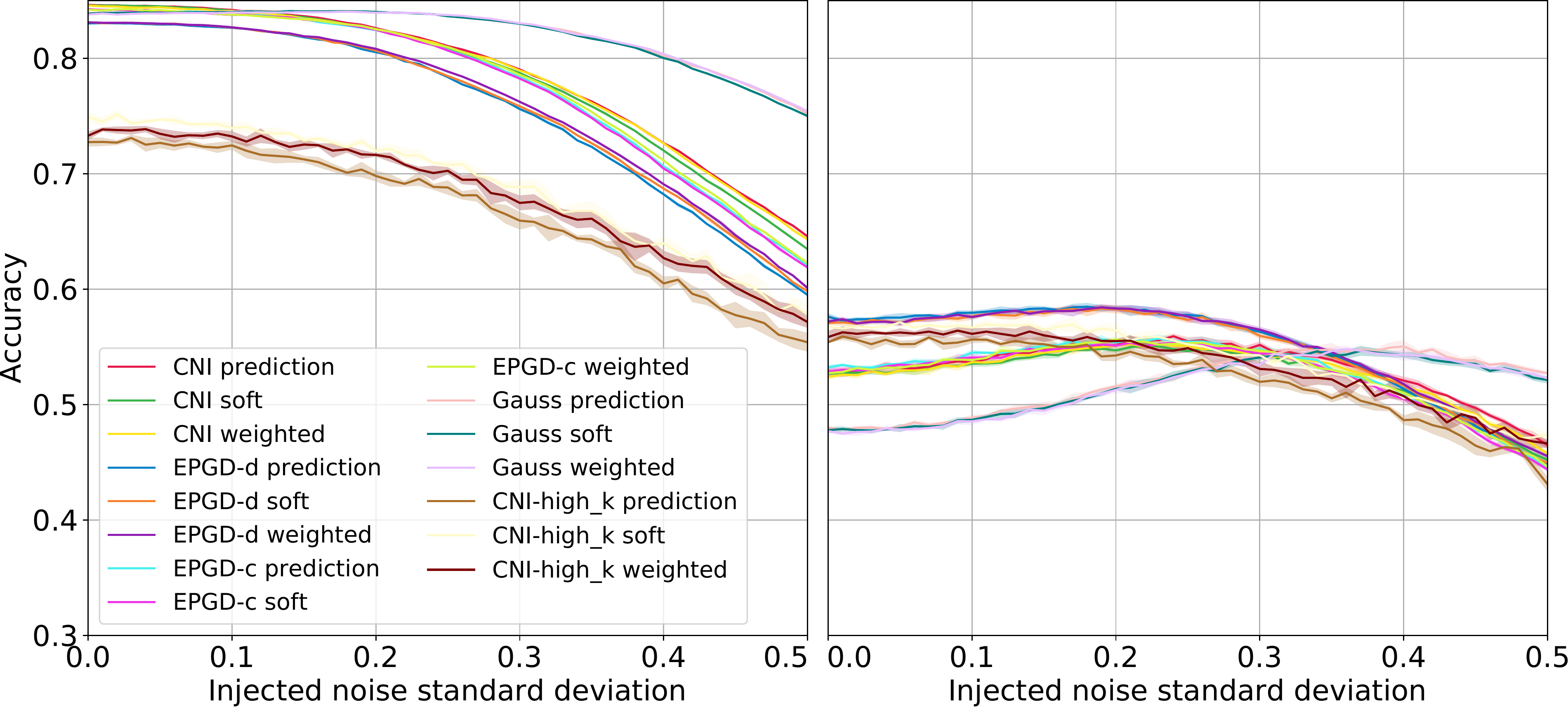}
 \caption{Accuracy as a function of injected noise strength with $M=8$ for all base model and all smoothing methods on clean data (left) and under PGD attack (right).}
\label{fig:all_8}
\end{figure*}

\begin{figure*}
 \centering
    \begin{subfigure}{0.49\linewidth}
        \includegraphics[width=\linewidth]{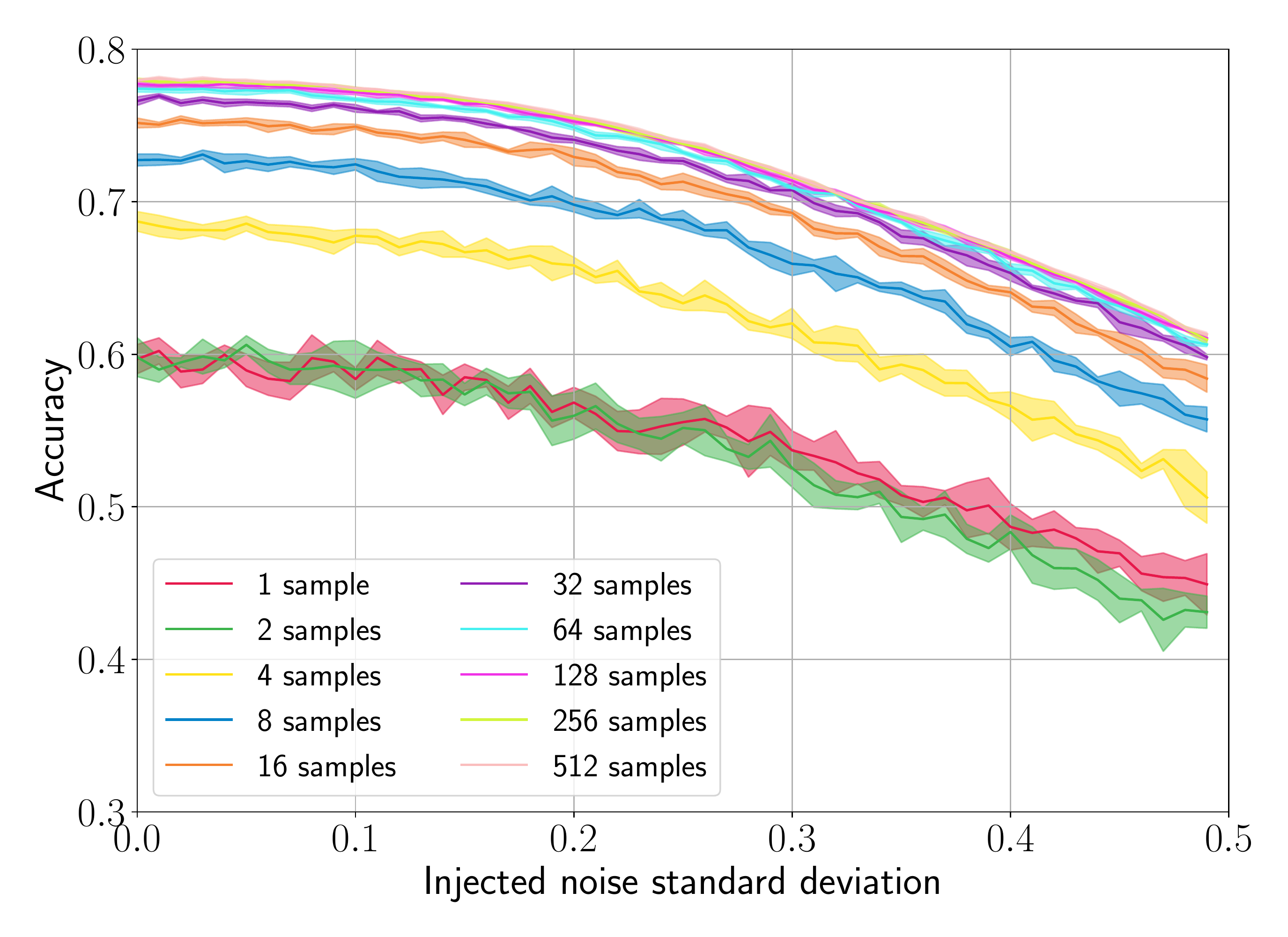}
        \subcaption{}
        \label{fig:pgd56_van}
    \end{subfigure}
    \hfill
    \begin{subfigure}{0.49\linewidth}
        \includegraphics[width=\linewidth]{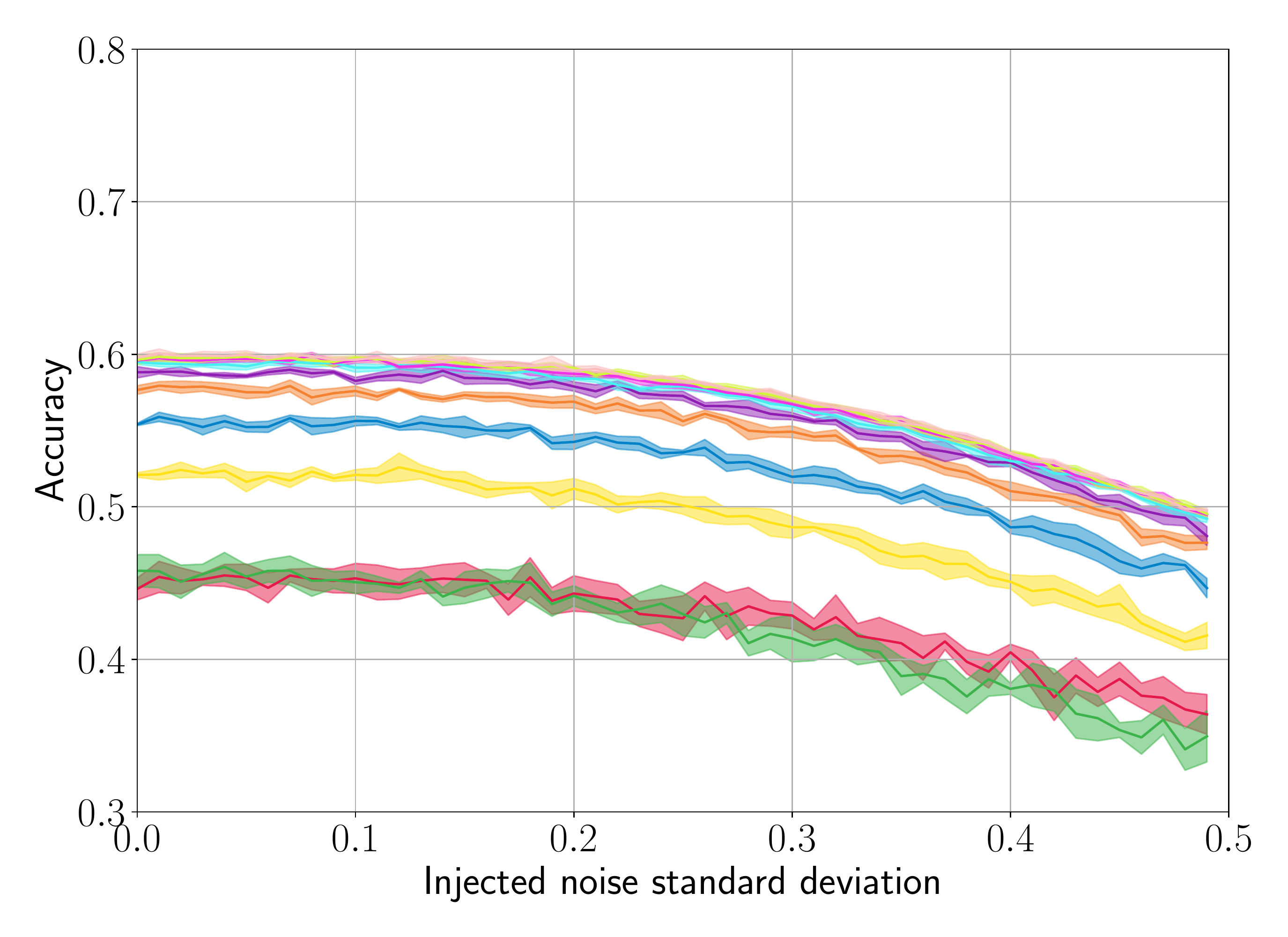}
        \subcaption{}
        \label{fig:pgd56_pgd}
    \end{subfigure}
 \caption{Accuracy as a function of injected noise strength with different numbers of samples for a CNI-W $(k=56)$ fine-tuned model with prediction smoothing on clean data (left) and under PGD attack (right). }
\label{fig:pgd56_model}
\end{figure*}

    \begin{figure}
        \includegraphics[width=\linewidth]{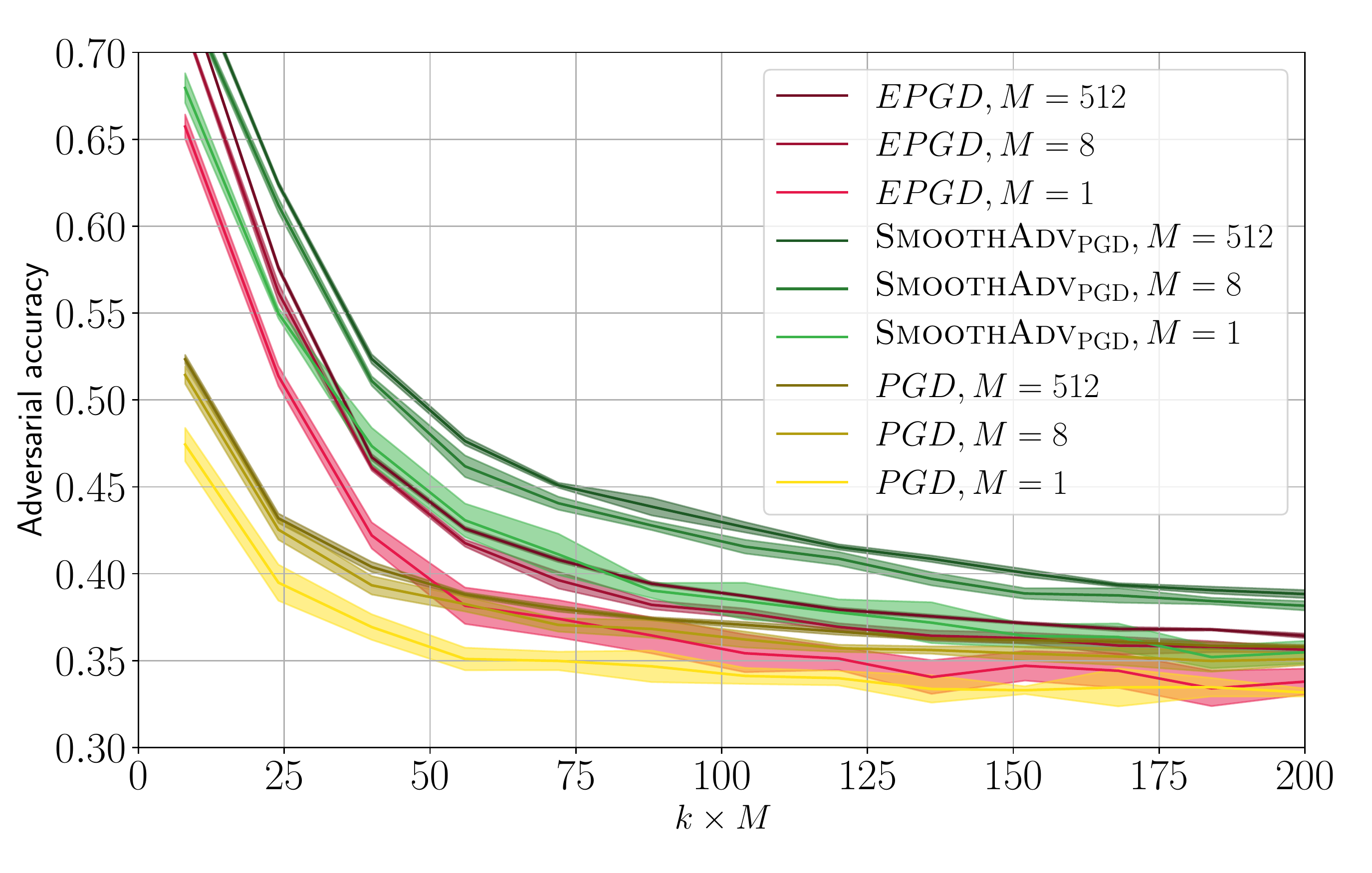}
    \caption{
    Accuracy of CNI-W model with prediction smoothing under PGD, EPGD and $SmoothAdv_{PGD}$ attacks with different iteration numbers ($k$).
    We used $M=1,8,512$ for smoothing, with a fixed noise standard deviation $\sigma=0.24$. EPGD and $SmoothAdv_{PGD}$ attacks averaged the gradients over $M_{\text{backward}}=8$ samples in each attack iteration.
    }
        \label{fig:km}
    \end{figure}

\cref{tab:Nattack_wideresnet} presents the additional evaluation of our method on a black box attack (NAttack \cite{li2019nattack}) compared with prior art. Our method substantially outperforms them.   
In addition, we evaluated the impact of noise injection levels on a PGD-56 pre-trained model for different numbers of samples. From \cref{fig:pgd56_model} we conclude that clean and adversarial accuracy reaches it nominal values for almost vanishing noise levels.

\cref{fig:all_8} demonstrates that there is a single optimal value of noise strength for each base model, independent of the smoothing method.
In all the figures we see a single maxima, which is fixed for each of the base models and unrelated to the smoothing method. This indicates that for each such model, there should be an optimal standard deviation of the injected noise.

In \cref{fig:km} we compared the effectiveness of the attacks compared to their computational complexity $k\times M$, for the PGD, EPGD and $SmoothAdv_{PGD}$ attacks.
PGD attacks are the most effective for lower $k\times M$, but converge to the same value as EPGD attacks without smoothing. In contrast, $SmoothAdv_{PGD}$ attacks both with and without smoothing are less effective for all values of $k\times M$  and converge to a higher accuracy in all cases. This could indicate that EPGD makes better use of the spatial information encoded in the multiple samples, so much so that it produces an attack comparable to PGD with $\nicefrac{1}{8}$ of the required iterations.
 Moreover, smoothing is more effective for EPGD and $SmoothAdv_{PGD}$, since both are using smoothing while constructing the attack. The accuracy under smoothing is better and converges to higher values for all attacks. This aids in validating the effectiveness of our method. 
 
\begin{figure*}
 \centering
 \includegraphics[width=\linewidth]{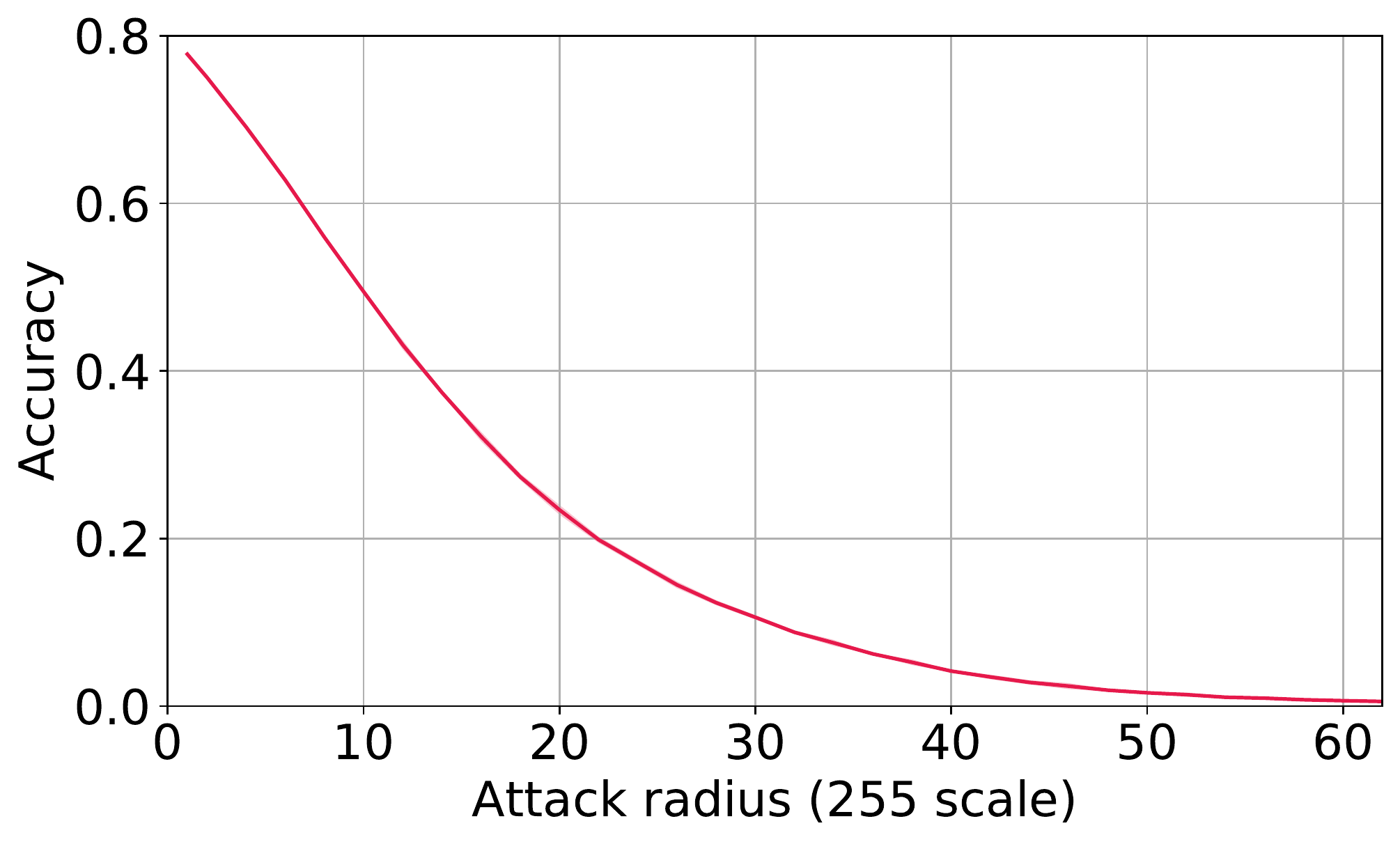}
\caption{Accuracy of the CNI-W model with prediction smoothing over 8 iterations under the PGD attack with different attack radii. Standard deviation is smaller than line width.}
        \label{fig:eps_ablation}
\end{figure*}

We evaluated our best-performing model against a PGD attack with  different strengths, $\epsilon$, to study the effect of transferring defense on attacks of different strengths. Results are shown in \cref{fig:eps_ablation}.

\begin{table}
    \centering
    \caption{
    Results of NAttack black-box attacks on our method applied on PNI-W and CNI-W with prediction smoothing, (a) ResNet-20 on CIFAR-10. (b) Wide-ResNet-34 on CIFAR-10.
    Our Smooth PNI-W and CNI-W methods uses $M=1,4,8$, $\sigma=0.24$. $^\dagger$ denotes our results based on code provided by the authors or our re-implementation.
     }
    \centering
    \begin{subfigure}{0.495\linewidth}
    \begin{tabular}{@{}cc@{}} 
        \toprule
         \textbf{Method} & \textbf{Accuracy, \%}  \\ 
        \midrule
        Adv. training \cite{madry2018towards}$^\dagger$  & $33.11$ \\
        Smooth PNI-W (our, $m=1$) &$47.17$\\
        Smooth PNI-W (our, $m=4$) &$50.29$\\
        Smooth PNI-W (our, $m=8$) &$50.78$\\
        Smooth CNI-W (our, $m=1$) & $48.83$  \\
        Smooth CNI-W (our, $m=4$) & $50.98$  \\
        Smooth CNI-W (our, $m=8$) & $\mathbf{51.56}$  \\
        \bottomrule
    \end{tabular}
    \subcaption{}
    \label{tab:Nattack_resnet}
    \end{subfigure}
    \hfill
    \begin{subfigure}{0.495\linewidth}
    \begin{tabular}{@{}cc@{}} 
        \toprule
         \textbf{Method} & \textbf{Accuracy, \%}  \\ 
        \midrule
        Adv. training \cite{madry2018towards}$^\dagger$  & $43.75$ \\
        Smooth PNI-W (our, $m=1$) &$\mathbf{56.25}$\\
        Smooth PNI-W (our, $m=4$) &$55.06$\\
        MART \cite{Wang2020Improving}$^+$ &$47.02$  \\

        \bottomrule
    \end{tabular}
    \subcaption{}

    \end{subfigure}
    \label{tab:Nattack_wideresnet}
\end{table}

\end{document}